\documentclass[letter]{article}

\usepackage{spconf,amsmath,graphicx}
\usepackage{times}
\usepackage{latexsym}
\usepackage{amsmath}
\usepackage{url}
\usepackage{subcaption}
\usepackage{graphicx}

\usepackage{caption}
\usepackage{multirow}
\usepackage{amssymb}
\usepackage{mathrsfs}
\usepackage{graphics}
\usepackage{graphicx}
\usepackage{ifsym}
\usepackage{amsmath}
\usepackage{subcaption}
\usepackage{booktabs}
\usepackage{multirow}
\usepackage{array}
\usepackage{bbm}
\usepackage{multicol}
\usepackage{blindtext}
\usepackage{algorithm}
\usepackage{algpseudocode}
\usepackage{tikz}

\usepackage{lipsum}
\usepackage{ntheorem}
\usepackage{makecell}

\let\oldbibliography\thebibliography
\renewcommand{\thebibliography}[1]{%
  \oldbibliography{#1}%
  \setlength{\itemsep}{0pt}%
}

\DeclareCaptionFont{10pt}{\fontsize{10pt}{12pt}\selectfont}
\title{\textbf{Neural Coreference Resolution based on Reinforcement Learning}}
\name{Yu Wang and Hongxia Jin}
\address{AI Center, Samsung Research America, Mountain View, CA, USA}

\begin{document}

\maketitle
\begin{abstract}
The target of a coreference resolution system is to cluster all mentions that refer to the same entity in a given context. All coreference resolution systems need to solve two subtasks; one task is to detect all of the potential mentions, and the other is to learn the linking of an antecedent for each possible mention. 
In this paper, we propose a reinforcement learning actor-critic-based neural coreference resolution system, which can achieve both mention detection and mention clustering by leveraging an actor-critic deep reinforcement learning technique and a joint training algorithm. We experiment on the BERT model to generate different input span representations. Our model with the BERT span representation achieves the state-of-the-art performance among the models on the CoNLL-2012 Shared Task English Test Set.
\end{abstract}
\section{Introduction}
A coreference resolution system normally needs to consider two tasks; one task is to identify the valid mentions appearing in a given context, and the other is to group mentions into different clusters such that mentions in each cluster point to the same target entity. The coreference systems can be either rule-based models \cite{lee2011stanford} or machine learning-based structures \cite{bjorkelund2014learning,lee2017end}. Most of these systems require complicated and hand-crafted heuristic features, which can be difficult to scale.

A recent trend is the use of neural-based coreference systems, which can avoid most of those hand-engineered features. One of the earliest works is the end-to-end neural coreference resolution system proposed by \cite{lee2017end}. There are two key scores in a neural coreference model; one is the mention score, and the other is the antecedent score. The two scores accordingly correspond to the mention detection subtask and mention resolution subtask. It is a common practice to calculate these two scores separately using decoupled models. In \cite{zhang2018neural}, however, a joint loss function is leveraged to concurrently optimize the performance of both tasks. Additionally, in the same work, a biaffine attention model is used to compute antecedent scores such that better performance can be obtained in comparison to the models using simple feed-forward networks.

Despite the decent performance exhibited by these neural coreference models, there are still many remaining challenges. One is that there may exist many varieties for a single entity's mentions, and it is very unlikely that our training data can explicitly label all of these singleton mentions. Exiting models suffer from mention proposal \cite{zhang2018neural}. Also most of the neural-based models are effective at identifying and clustering mentions in a given sentence context but perform worse once the mention's sentence context changes. This is primarily because most models are trained using each sentence or document as a sample and attempt to capture the relations between mentions within each given training sample. However, models will not function effectively if the mention appears in another new sentence or document in which its context is totally different.

To this end, we propose a new actor-critic deep reinforcement learning-based neural coreference resolver, to better handle the mention varieties by considering the mention-level training samples instead of only the sentence-level or document-level samples. This proposed algorithm gives us the advantages that it can significantly alleviate the negative effect from the noisy sentence-level information, but still can keep the necessary contextual cues when generating the mention representations. Furthermore, we also introduce a distance based reward function such that it can take the distance between two mentions into consideration since co-reference is sensitive to the mention distance. The model can also better handle the mention's stochasticity in the training data (\emph{i.e.} the same mentions appear in different document contexts) by leveraging the actor-critic-based deep reinforcement learning technique. Furthermore, we propose an end-to-end system, which can jointly achieve mention detection and mention clustering by using an augmented loss function.

We evaluate the system on the CoNLL-2012 English dataset and achieve the new state-of-the-art performance of 87.5$\%$ average F1 score by using this new model with BERT pre-trained representations.
\section{System Structure}
In this section, we explain the detailed structure of the new proposed reinforcement learning actor-critic-based neural coreference resolver (AC-NCR). However, before we do that, we first show how to generate the span representations in our system. We will also show how to train the model using the actor-critic deep reinforcement learning model, as well as jointly training mention clustering and mention detection using the augmented actor-critic loss functions.
\subsection{Span Representation}
In this section, we present our span representations by using the BERT based model.
Figure \ref{span2} shows the model structure used to generate span representations $m_i$ ($0 \leq i \leq k$) for all valid mention entities, each mention entity may contain one or several tokens $w_t$. For each token $w_t$ in the sentence, its embedding $o_t$ is generated by the BERT embedding layer. The generated embeddings are then passed to an attention layer with their outputs as the head-finding attention vectors $w_i^{head}$ for the mention spans:

\begin{figure}[ht!]
  \centering
  \includegraphics[width=1\linewidth]{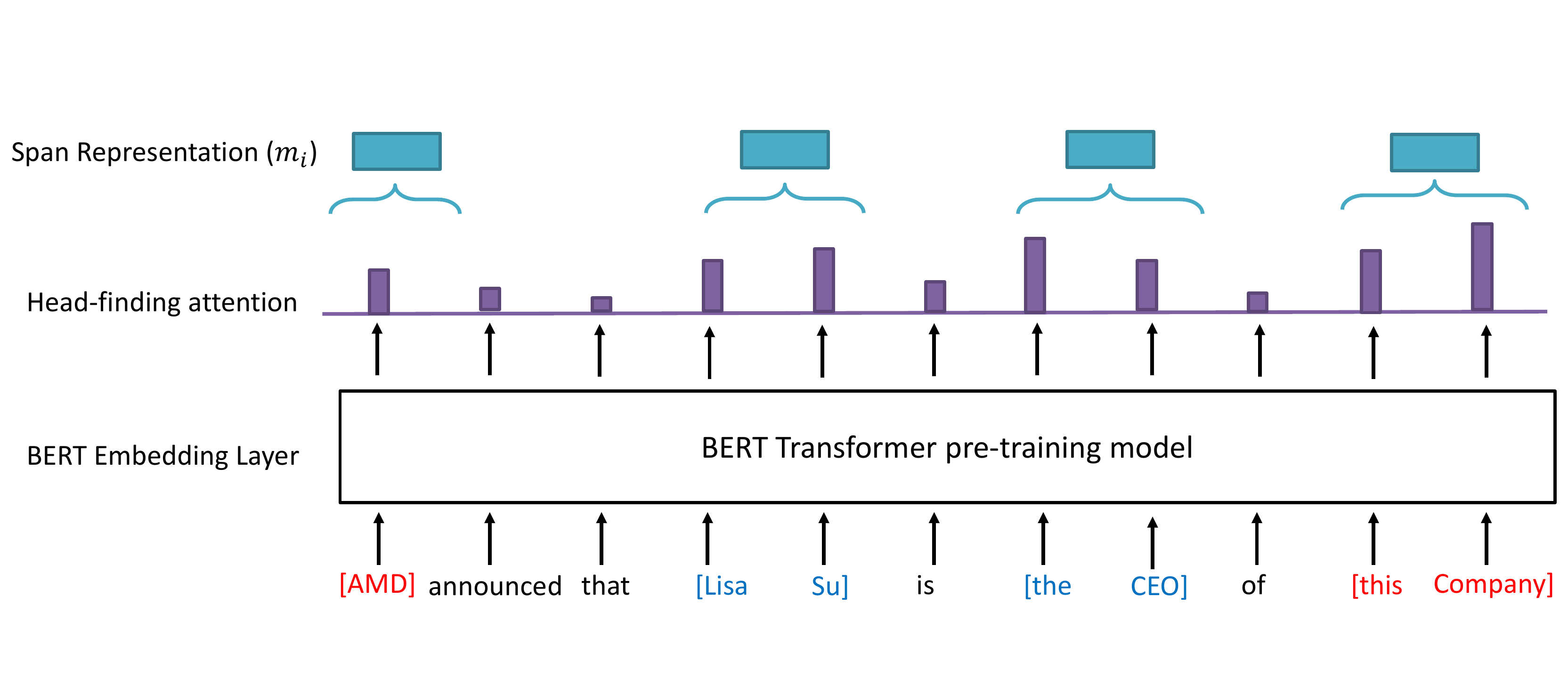}
  \caption{The model structure for generating a BERT Transformer span representations}
  \label{span2}
\end{figure}
\begin{equation}
\begin{split}
o_{t}&={v}_{o}^Tf_{bert}(w_t)\\
{\alpha}_{i,t}&=\dfrac{e^{o_t}}{\sum_{k=i_{start}}^{i_{end}}e^{o_k}}\\
w_i^{head}&=\sum_{k=i_{start}}^{i_{end}}{\alpha}_{i,t}w_t
\end{split}
\label{spanrepresentation}
\end{equation}
where ${\alpha}_{i,t}$ is the word-level attention parameter for the $t^{th}$ word in the $i^{th}$ mention, and $w_i^{head}$ is the head-finding attention vector.

$i_{start}$ stands for the starting word position in the $i^{th}$ mention, and correspondingly $i_{end}$ is the ending word position. 

Inspired by a similar setup as in \cite{zhang2018neural}, the span presentation is a concatenation of four vectors, which is defined as:
\begin{equation}
m_i=o_{i_{start}}\oplus o_{i_{end}}\oplus w_i^{head} \oplus \lambda_{i}
\label{spanrepresentation2}
\end{equation}

Since BERT is a pre-trained language model, our training data sample needs to be split into multiple segments, each of which has a maximum segment length. In this paper, we test on different maximum segment lengths of 128, 256 and 512, and the length of 128 gives us the best performance. Hence, for the rest of this paper, we use 128 as the maximum segment length for training our BERT based model.
\subsection{AC-NCR Model Design}
\begin{figure*}[ht!]
  \centering
  \includegraphics[width=0.68\linewidth]{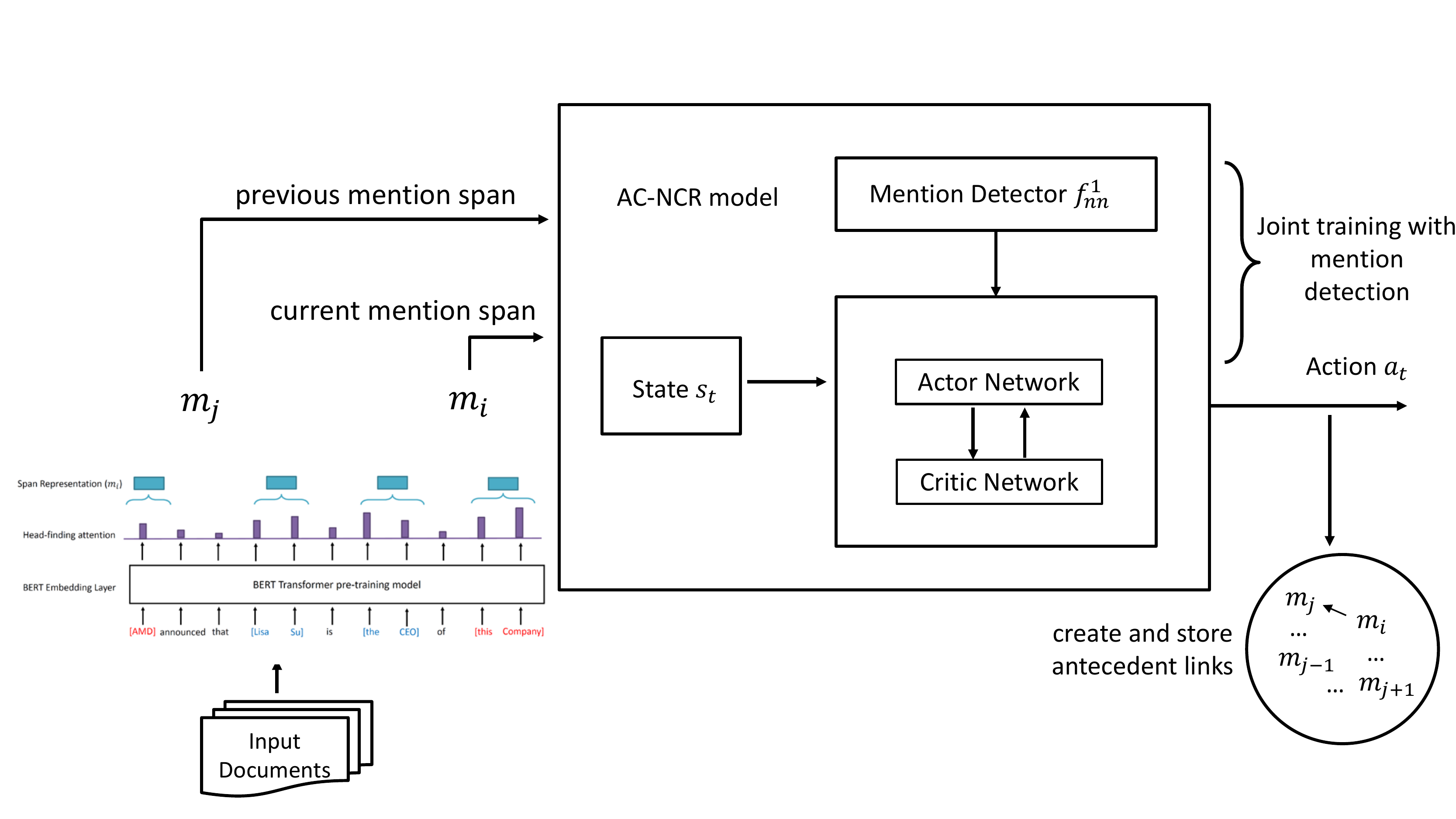}
  \caption{Model structure of AC-NCR: the model takes two mentions $[m_i,m_j]$ as its input at each round, and all text spanning up to $K=10$ words is considered.}
  \label{training}
\end{figure*}

In this section, the design details of an actor-critic-based neural coreference resolver (AC-NCR) are given. The entire system structure is proposed as in Figure \ref{training}.

The AC-NCR model is trained based on deep reinforcement learning (DRL) technique. Deep reinforcement learning has been widely used in a variety of NLP and machine learning tasks \cite{wang2018deep,wang2018boosting,wang2019deep,narendra2016fast,wang2019neural,wang2020actor,wang2021coarse}. The input to the model is a state $s_t$, and its output is an action $a_t$. The model is trained with the target to maximize the expected reward generated by the reward function $r_t$. Their definitions are as given below:

{\bf{State:}}The state is defined by concatenating the current mention span representation $m_i$ with one of its previous spans $m_j$ ($1\leq j \leq i-1$). It contains the information of both spans to decide whether or not $m_j$ is an antecedent of $m_i$.

{\bf{Action:}}
There are three different types of actions defined in this system. The first one is to move the current span $m_i$ by one step to the right, \emph{i.e.} $m_i \rightarrow m_{i+1}$, which indicates that $m_i$ has found an antecedent $m_j$, and hence moves to the next span. At the same time, the second term representing the antecedent span is reset to $m_0$, and the state $s_t=[m_i,m_j$] is stored as $m_j$, which is an antecedent of $m_i$ formed by linking the two mentions. On the other hand, if the previous span $m_j$ is not an antecedent of $m_i$, the system should move to $m_j$'s next span $m_{j+1}$ for further evaluation. There is a third action: if input $s_t=[m_i$,$m_j$] satisfies the condition that $i=j+1$, \emph{i.e.} $m_i$ is the next mention of $m_j$ in the context, it is possible that there is no antecedent for $m_i$, and action $a_t$ is to transfer the system from current state $s_t=[m_i,m_j]$ to the new state $s_{t+1}=[m_{i+1},m_0]$, but the antecedent link is not stored as action 1. To represent these three actions mathematically, we have the following:

\begin{equation}
a_t= \begin{cases}
               [m_i,m_j]\rightarrow[m_{i+1},m_0],\text{store }[m_i,m_j]\\
               [m_i,m_j]\rightarrow[m_{i},m_{j+1}]\\
               [m_i,m_j]\rightarrow[m_{i+1},m_{0}],\text{don't store}
            \end{cases}
\label{action}
\end{equation} 
The state $s_t$ will then transition to $s_{t+1}$, directed by action $a_t$. 

%
 
{\bf{Reward:}}
Inspired by \cite{dozat2016deep} and \cite{zhang2018neural}, the reward $r_t$ at a state $s_t$ is defined using a biaffine attention technique to model the likelihood of $m_j$ as an antecedent of $m_i$, by jointly considering the distance between two mentions. The reward is hence defined as:
\begin{equation}
\begin{split}
r_t &= e^{-\gamma \parallel i-j \parallel}(v_m^Tf_{nn}^1(m_i)+v_m^Tf_{nn}^1(m_j)\\
&+f_{nn}^2(m_j)U_{bi}f_{nn}^2(m_i)+v_{bi}^Tf_{nn}^3(m_i))
\end{split}
\label{reward}
\end{equation} 
where $f_{nn}^{1,2,3}(\cdot)$ are feedforward networks to reduce span representation dimensions, ${U}_{bi} \in \mathbb{R}^{k\times k}$ and ${v}_{bi} \in \mathbb{R}^{k\times 1}$ are the linear transformation matrix/vector, and $k$ is the dimension of the output of $f_{nn}(\cdot)$, $\gamma \in (0,1)$ is the user chosen decay factor. The exponential decay term $e^{-\gamma \parallel i-j \parallel}$ indicates that a smaller reward is assigned when the distance between two mentions ($m_i$,$m_j$) (\emph{i.e.} $\parallel i-j \parallel$) becomes larger. The reward design based on two mentions' distance follows a straight-forward observation that: the probability of two mentions are coreferent to each other becomes smaller when distance between them becomes larger.

As the reward defined in \ref{reward}, the terms $v_m^Tf_{nn}^1(m_i)$ and $v_m^Tf_{nn}^1(m_j)$ are the mention scores to measure their likelihoods as entity mentions. $f_{nn}^2(m_j){U}_{bi}f_{nn}^2(m_i)$ defines the compatibility between $m_i$ and $m_j$, and ${v}_{bi}^Tf_{nn}^3(m_i)$ measures the likelihood of the current mention $m_i$ having an antecedent.
\section{Model Training}
\subsection{Actor-Critic Model Training}
As mentioned previously, in this DRL system, we use an actor-critic method-based DRL model, where two neural networks are trained to model the actor and critic separately. One important reason is that we use an actor-critic model instead of other DRL models is because it is known to converge more smoothly and have better training performance on a system with a large state space \cite{lillicrap2015continuous}, which is like our scenario.

%

There are two loss functions corresponding to the actor and critic networks, specified as:
\begin{equation}
\begin{split}
\mathcal{L}_{actor}&=-\log \pi_{\theta} (a_t|s_t)(r_t+\gamma V(s_{t+1})-V(s_t))\\
\mathcal{L}_{critic}&=(r_t+\gamma V(s_{t+1})-V(s_t))^2
\end{split}
\end{equation}
where $\gamma$ is a discount factor, and $\pi_{\theta}$ is the policy probability of taking action $a_t$. 
%
\subsection{Joint Training with Mention Detection}
Our AC-NCR model performs mention clustering once possible mentions are detected. The model relies on a feed-forward neural network mention classifier $f_{nn}^1(m_i)$ to generate the mention score to measure its likelihood as an entity's mention. During training, only mention cluster labels are available, rather than antecedent links, and hence we propose two augmented joint loss functions for training our AC-NCR model by also taking the mention detection loss into consideration. Before that, we must first define the mention detection loss as:
\begin{equation}
\begin{split}
\mathcal{L}_{detect}&= y(m_i)\log(S(m_i))\\
&+(1-y(m_i))\log(1-S(m_i))
\end{split}
\end{equation}
where $y(m_i)$ is equal to 1 if $m_i$ is in one of the gold mention clusters, and otherwise is equal to 0. $S(m_i)$ is a sigmoid function of $m_i$.

Since mention detection is a prerequisite for both actor and critic network, both of their loss functions are augmented by the detection loss $\mathcal{L}_{detect}$ as:
\begin{equation}
\begin{split}
\mathcal{L}_{actor}^{\prime}&=\mathcal{L}_{actor}+\mathcal{L}_{detect}\\
\mathcal{L}_{critic}^{\prime}&=\mathcal{L}_{critic}+\mathcal{L}_{detect}
\end{split}
\label{lossac}
\end{equation}

{\it{Remarks:}} Mention detection can also treated as a separate semantic parsing task, and can be achieved by generic tagging and parsing models for natural language understanding as in \cite{wang2018bi,wang2018new,wang2018neural,wang2019deep,wang2020interactive,wang2020bi,wang2020new,wang2020multi,shen2021system,wang2021adversarial}. In this section, we use a joint loss to train the mention detection model such that it can leverage the extra information provided by the clustering model.
 
\begin{table*}[t]\scriptsize
\parbox{1\linewidth}{
\centering
	\caption{Experimental results on CoNLL-2012 English test set}
	\label{table:comparisonCoNLL}
	\begin{tabular}{>{\centering\arraybackslash}p{4cm}|>{\centering\arraybackslash}p{0.5cm}>{\centering\arraybackslash}p{0.5cm}>{\centering\arraybackslash}p{0.5cm}>{\centering\arraybackslash}p{0.4cm}>{\centering\arraybackslash}p{0.5cm}>{\centering\arraybackslash}p{0.5cm}>{\centering\arraybackslash}p{0.5cm}>{\centering\arraybackslash}p{0.5cm}>{\centering\arraybackslash}p{0.5cm}>{\centering\arraybackslash}p{0.6cm}}
		\toprule
		\multirow{1}{*}{\textbf{}} & \multirow{1}{*}{\makecell{\textbf{}}} & \multirow{1}{*}{\makecell{\textbf{$MUC$}}}& \multirow{1}{*}{\makecell{\textbf{}}}& \multirow{1}{*}{\makecell{\textbf{}}} & \multirow{1}{*}{\makecell{\textbf{$B^3$}}}& \multirow{1}{*}{\makecell{\textbf{}}}& \multirow{1}{*}{\makecell{\textbf{}}} & \multirow{1}{*}{\makecell{\textbf{$CEAF_{\phi_4}$}}}& \multirow{1}{*}{\makecell{\textbf{}}}& \multirow{1}{*}{\makecell{\textbf{}}}\\		
		\midrule
		\multirow{1}{*}{\textbf{Model}} & \multirow{1}{*}{\makecell{\textbf{P}}} & \multirow{1}{*}{\makecell{\textbf{R}}}& \multirow{1}{*}{\makecell{\textbf{F1}}}& \multirow{1}{*}{\makecell{\textbf{P}}} & \multirow{1}{*}{\makecell{\textbf{R}}}& \multirow{1}{*}{\makecell{\textbf{F1}}}& \multirow{1}{*}{\makecell{\textbf{P}}} & \multirow{1}{*}{\makecell{\textbf{R}}}& \multirow{1}{*}{\makecell{\textbf{F1}}}& \multirow{1}{*}{\makecell{\textbf{Avg. F1}}}\\
\midrule
		\multirow{2}{*} {}{\textbf{AC-NCR w BERT}}  & 93.6 &90.4 &92.5 &86.0 &85.3&85.9 &84.2 &82.1 &84.1 &{\bf{87.5}}\\
		\midrule
		
		\multirow{2}{*} {}\cite{wu2020corefqa}  & 88.6 &87.4 &88.0 &82.4 &82.0 &82.2 &79.9 &78.3 &79.1 &83.1\\
		\multirow{2}{*} {}\cite{joshi2020spanbert}  & 85.8 &84.8 &85.3 &78.3 &77.9 &78.1 &76.4 &74.2 &76.3 &79.6\\
		\multirow{2}{*} {}\cite{joshi2019bert}  & 84.7 &82.4 &83.5 &76.5 &74.0 &75.3 &74.1 &69.8 &71.9 &76.9\\
				
		\multirow{2}{*} {}\cite{kantor2019coreference}  & 82.6 &84.1 &83.4 &73.3 &76.2 &74.7 &72.4 &71.1 &71.8 &76.6\\ 
		
		\multirow{2}{*} {}\cite{fei2019end}  & 85.4 &77.9 &81.4 &77.9 &66.4 &71.7 &70.6 &66.3 &68.4 &73.8\\ 
				
		\multirow{2}{*} {}\cite{lee2018higher}  & 81.4 &79.5 &80.4 &72.2 &69.5 &70.8 &68.2 &67.1 &67.6 &73.0\\ 
		
		\multirow{2}{*} {}\cite{zhang2018neural}  & 79.4 &73.8 &76.5 &69.0 &62.3 &65.5 &64.9 &58.3 &61.4 &67.8\\ 
		
		\multirow{2}{*}{}\cite{lee2017end}  &78.4 &73.4 &75.8 &68.6 &61.8 &65.0 &62.7 &59.0 &60.8 & 67.2  \\
		\bottomrule		
	\end{tabular}
}
\end{table*}

\begin{table}[h]\scriptsize
\parbox{1\linewidth}{
\centering
	\caption{Ablation Study on Models with/without Mention Detection Loss}
	\label{table:ablationstudy}
	\begin{tabular}{>{\centering\arraybackslash}p{5cm}|>{\centering\arraybackslash}p{1cm}}
		\toprule
		\multirow{1}{*}{\textbf{Model}} & \multirow{1}{*}{\makecell{\textbf{Avg. F1}}}\\
		\midrule
		\multirow{2}{*}{}AC-NCR w BERT & 87.5\\ 	
		\multirow{2}{*}{}AC-NCR w BERT without mention detection  &85.1\\
		\midrule		
		\multirow{2}{*}{}\cite{wu2020corefqa}  &83.1\\
		\bottomrule		
	\end{tabular}
}
\end{table}

\section{Experiment}
\subsection{Dataset}
The dataset we used for experimentation is the CoNLL-2012 Shared Task English data \cite{pradhan2012conll} based on the OntonNotes corpus. The training set contains 2,802 documents, and the validation set and test set contain 343 and 348 documents, respectively. Similar to the previous works, we use three different metrics: MUC \cite{vilain1995model}, B$^3$ \cite{bagga1998algorithms} and CEAF$_{\phi_4}$ \cite{luo2005coreference}, and report the respective precision, recall and F1 scores.

\subsection{Model Implementation}
In this paper, we consider all spans up to 250 antecedents and 10 words.
In our BERT setup, we use 128 as the maximum segment length for training our BERT model. Both the actor and critic networks use the LSTM structure with hidden layer size of 200. The feedforward neural networks used to generate mention detection scores include two hidden layers with 150 units and ReLU activations. To be comparable with previous popular models, we include features (speaker ID, document genre, span distance and span width) as the 20-dimensional learned embeddings. For the $f_{nn}^{(1,2,3)}$, we use a dropout rate of 0.5.

\begin{table}[h]\scriptsize
\parbox{1\linewidth}{
\centering
	\caption{Ablation Study between the Model with an Actor-Critic DRL structure and the Model with a supervised trained BERT model}
	\label{table:ablationstudy2}
	\begin{tabular}{>{\centering\arraybackslash}p{5cm}|>{\centering\arraybackslash}p{1cm}}
		\toprule
		\multirow{1}{*}{\textbf{Model}} & \multirow{1}{*}{\makecell{\textbf{Avg. F1}}}\\
		\midrule	
		\multirow{2}{*}{}AC-NCR w BERT & 87.5\\ 	
		\multirow{2}{*}{}Supervised LSTM w BERT &78.6\\
		\midrule		
		\multirow{2}{*}{}\cite{wu2020corefqa}  &83.1\\
		\bottomrule		
	\end{tabular}
}
\end{table}

\begin{table}[h]\scriptsize
\parbox{1\linewidth}{
\centering
	\caption{Mention detection subtask on development set}
	\label{table:mentiondetection}
	\begin{tabular}{>{\centering\arraybackslash}p{2.5cm}|>{\centering\arraybackslash}p{1cm}>{\centering\arraybackslash}p{1cm}>{\centering\arraybackslash}p{1cm}>{\centering\arraybackslash}p{1cm}}
		\toprule
		\multirow{1}{*}{\textbf{Model/Span Width}} & \multirow{1}{*}{\makecell{\textbf{ 1-2}}}& \multirow{1}{*}{\makecell{\textbf{ 3-4}}}& \multirow{1}{*}{\makecell{\textbf{5-7}}}& \multirow{1}{*}{\makecell{\textbf{8-10}}}\\
		\midrule
		\multirow{2}{*}{}AC-NCR w BERT & 92.8& 83.4& 72.8& 65.6\\ 	
		\midrule		
		\multirow{2}{*}{}\cite{joshi2020spanbert}  &87.6& 76.4& 63.4& 52.8\\
		\bottomrule		
	\end{tabular}
}
\end{table}
\subsection{Performance}
In Table \ref{table:comparisonCoNLL}, we compare our model with the previous state-of-the-art models. Here, we only compare single models without any ensemble for fairness. It can be observed that our model with BERT pre-training embedding achieves the best performance in terms of all metrics. Especially compared to the model with a similar biaffine technique, but being trained in a supervised manner instead,our AC-NCR model generates a higher recall value since the DRL training technique can cover more stochastic mention patterns or antecedent links, which are not exhibited in the training data.
\subsection{Ablation Study}
\subsubsection{Effect of Joint Training with Mention Detection}
To evaluate the impact of the mention detection upon the mention cluster task, we remove the mention detection term in the loss function given in \ref{lossac}. The result is given in Table \ref{table:ablationstudy}. We can observe that the overall F1 score decreases by 2.4 on AC-NCR with BERT. The drop is mainly because the model does not leverage any useful information from mention detection in this setup.
\subsubsection{Effect of Actor-Critic DRL model}
In order to evaluate the impact of our actor-crtic DRL model over the task, we replace our DRL model structure by a supervised a supervised LSTM structure with a hidden layer size of 200. The embeddings are also generated by the BERT model. The feedforward neural networks used to generate mention detection scores are kept as the same network with 150 units and ReLU activations. In order to train the model in a supervised manner, we use the reward $r_t$ as the coreference score or the training label for each given mention pairs fed into the system. The inputs of the training system are the mention representations of a sentence sample as in \cite{zhang2018neural}. The mentions score function and loss functions are used the same as in our AC-NCR setup. During inference, mentions are clustered by using the generated coreference score in a given sample.

The results given in Table \ref{table:ablationstudy2} shows that the model trained using actor-critic network structure performs far better than the same model trained supervised without using the DRL structure. One main reason is that the supervised trained model is not able to capture the mentions' stochasticity, especially when the same mentions appear in different contexts.
\subsection{Mention Detection Task}
To further understand our model, we separate the mention detection task from the joint task. In this setup, we consider spans with mention scores higher than zero as mentions. The mention detection accuracy versus span length is as shown in Table \ref{table:mentiondetection}. Our model with different embeddings both perform better than the current state-of-the-art model on the CoNLL-2012 dataset in terms of mention detection accuracy. It is also observed that the advantage becomes larger when the span width increases.
\section{Conclusion}
In this paper, we propose an end-to-end actor-critic-based neural coreference resolution system which can perform both mention detection and clustering jointly.  Our model with the BERT transformer span representation achieves the state-of-the-art performance among the models on the CoNLL-2012 Shared Task English Test Set.

\bibliographystyle{IEEEtran}

\bibliography{acncr}

\end{document}